\documentclass[runningheads]{llncs}

\usepackage{pifont}

\usepackage{mathtools} 
\usepackage{csquotes}
\usepackage{multirow}
\usepackage{longtable}
\usepackage[linesnumbered,noend,ruled,resetcount]{algorithm2e}
\usepackage{array}
\newcolumntype{C}[1]{>{\centering\let\newline\\\arraybackslash\hspace{0pt}}m{#1}}
\usepackage{graphicx}
\usepackage{rotating}

\begin{document}
\title{Impact of different belief facets on agents' decision --- a refined cognitive architecture to model the interaction between organisations' institutional characteristics and agents' behaviour}
\titlerunning{Impact of different belief facets on agents' decision}
%
\author{Amir Hosein Afshar Sedigh\inst{1}\and Martin K. Purvis\inst{1} \and Bastin Tony Roy Savarimuthu\inst{1}\and Christopher K. Frantz\inst{2}\and Maryam A. Purvis\inst{1}}
\authorrunning{A. H. Afshar Sedigh et al.}
%
\institute{Department of Information Science, University of Otago, Dunedin, New Zealand \and Department of Computer Science, Norwegian University of Science and Technology, Ametyst-bygget, A205, Gj{\o}vik, Norway. \email{amir.afshar@postgrad.otago.ac.nz}\\ \email{\{martin.purvis,tony.savarimuthu,maryam.purvis\}@otago.ac.nz}\\ \email{christopher.frantz@ntnu.no}
}

\maketitle              
\begin{abstract}
This paper presents a conceptual refinement of agent cognitive architecture inspired from the beliefs-desires-intentions (BDI) and the theory of planned behaviour (TPB) models, with an emphasis on different belief facets. This enables us to investigate the impact of personality and the way that an agent weights its internal beliefs and social sanctions on an agent's actions. The study also uses the concept of cognitive dissonance associated with the fairness of institutions to investigate the agents' behaviour. To showcase our model, we  simulate two historical long-distance trading societies, namely Armenian merchants of New-Julfa and the English East India Company. The results demonstrate the importance of internal beliefs of agents as a pivotal aspect for following institutional rules. 
\keywords{Institutions  \and BDI \and Agent-based simulation \and Facets of belief \and Cognitive dissonance.
}
\end{abstract}
\section{Introduction}

This paper extends a mental architecture to model dynamics in an agent's cognition. In other words, this paper presents a cognitive architecture inspired from the belief-desire-intention model (BDI) \cite{Bratman1988} and the theory of planned behaviour (TPB) \cite{fishbein2011predicting} to investigate agents' interactions with institutions.  

 In this study, we use agents to divide a complex social system into smaller action components. Also, an agent impacts the system through its decisions. A challenge in agent-based simulation is modelling the agents’ decision-making process (i.e. action deliberation) \cite{balke2014}. A class of studies addresses this challenge by employing the beliefs-desires-intentions (BDI) cognitive architecture. Some well-known extensions of the BDI cognitive architecture include the BOID \cite{BOID}, EBDI \cite{Pereira2008} and the BRIDGE \cite{Dignum2009}. Also, a formalisation of BDI (n-BDI) was presented to take into consideration the norms and their internalisation \cite{Criado2010}. Some researchers employed the BDI architecture to model agents' cooperation in institutionalised multi-agent systems \cite{BDIInst,Balke2012}. In some studies, different components of information are considered in modelling agent architecture \cite{Castelfranchi2000}. However, the stated study distinguishes the stored information based on their types and does not address the agents' interpretations given their enforcing meta-roles. 
 
 To our knowledge, none of the works in the area have taken into account a multi-faceted characterisation of beliefs as behavioural moderators, so as to produce nuanced behavioural outcomes that respond to the context and embedding of the agents therein. In this study, we are thus inspired from the BDI cognitive architecture and different belief facets of TPB. Also, we model the impact of different belief facets on agent decisions by drawing on extant social-psychological theories. More precisely, we model how institutional characteristics, including the fairness of institutions, impact agents' decisions. The impact of fairness on agents' decisions was studied by several researchers \cite{PahlWostl2004,logist2006,Levy2018}, in terms of fair behaviour of agents. In the stated studies the fairness concerned how individual agents punish each other, not the organisation (i.e.~the focus of this paper).  

Before providing the paper's organisation, we need to clarify that the nature of simulation and case studies contain some specification. Therefore, the simulation section cannot use the agent architecture as one may expect. The rest of this paper is organised as follows. Section \ref{cogarc} discusses our cognitive architecture. Section \ref{opersec} discusses operationalisation of the central aspects of the model. Section \ref{SimulationAndPArameters} sketches a comparative overview of two historical cases, along with a simulation model for those societies that incorporates the proposed architecture. Section \ref{results} presents the simulation results for the two societies. Section \ref{conc} discusses the findings and provides concluding remarks.

 
 \section{An overview of the cognitive architecture}
\label{cogarc}
\begin{figure}[hbt!]
		\centering
			\includegraphics[trim=1cm 18cm 1.6cm 0cm,clip, width=1\textwidth]{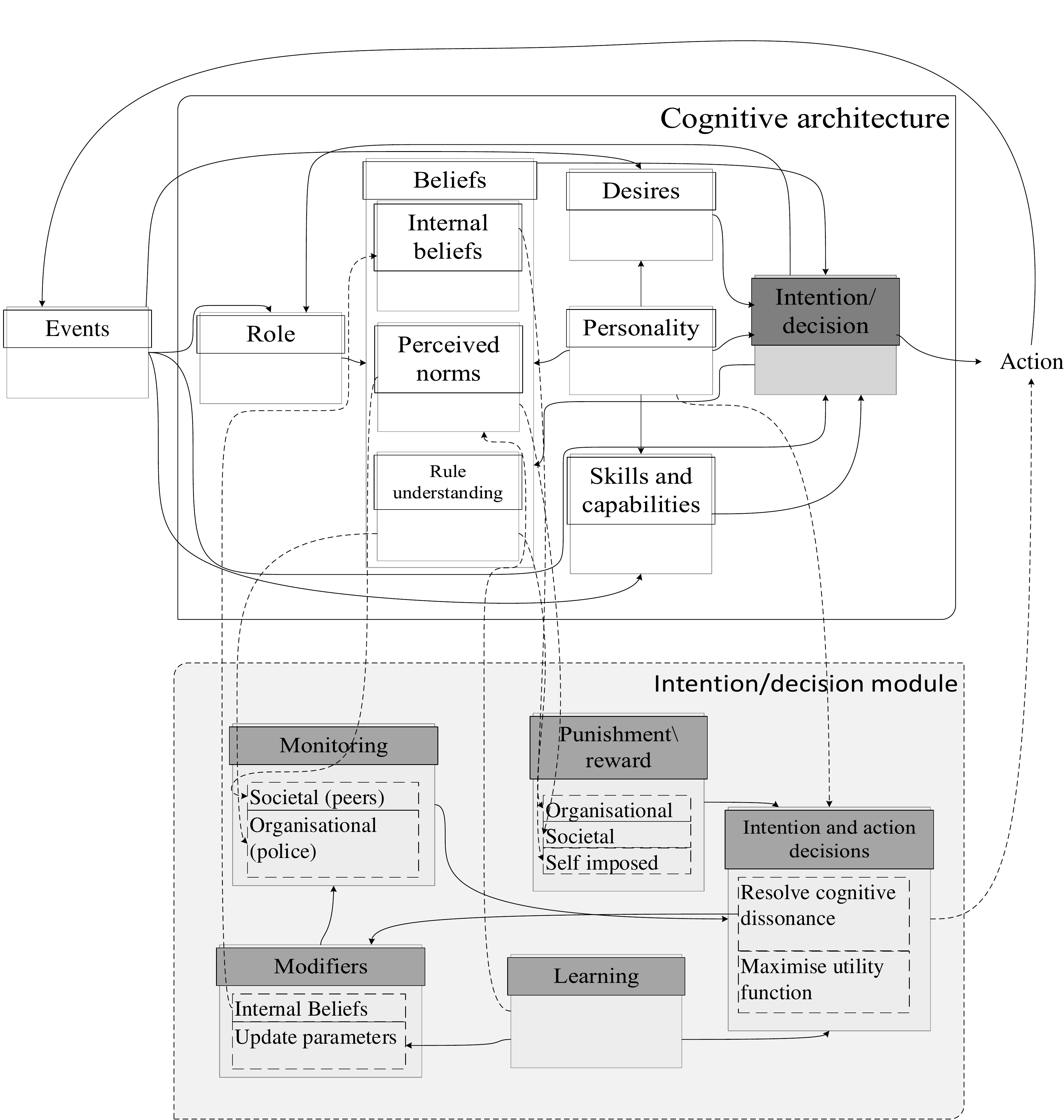}
		\caption{A schematic diagram of the proposed cognitive architecture. The intention/decision module of the cognitive architecture is expanded in the dashed box (bottom block). The relationships between the modules of the cognitive architecture and the sub-modules of the ``intention/decision module'' is given by dashed-lines. }
		\label{architec}
	\end{figure}

Our cognitive model inspired by the BDI and TPB models is depicted in Figure \ref{architec}. It includes a decision-making module that is expanded in the same figure in grey. What follows discusses different modules and their connections with one another. The architecture separates inputs, here referred to as ``Events'' that are external to the reasoning process from the actual cognitive architecture that operates on these inputs. The Cognitive architecture block represents an agent's cognitive decision-making components. Any action performed by an agent in a given iteration will be an input event for the agent itself and other agents in its social environment in the next iteration. For instance, an agent's cheating action in time $t$ will be an action recognised by associated agents for the next period ($t + 1$). In the following, we will discuss the internal components and underlying assumptions of the architecture in more detail.

What follows discusses different modules and their connections with one another. As can be seen, there are two separate blocks, namely a left block called `\textit{Events}' and a right block called `{Cognitive architecture}'. The {Events} block represents the events from an outside environment. The \textit{Cognitive architecture} block represents an agent's cognitive decision-making components. Note that when an action is performed by an agent, it will be an event for agents in the next iteration. 
What follows discusses the modules of cognitive architecture.
	

	\begin{itemize}
 \item \textbf{Roles:} An agent has a set of roles in society that indicate how a given rule impacts it \cite{Purvis2014}.
 \item \textbf{Beliefs:} To model beliefs, we are inspired by idea of different belief facets \cite{fishbein2011predicting}, such as Fishbein and Ajzen’s Reasoned Action/Planned Behaviour approach, but propose a variation of the belief facets. We consider three different punitive/rewarding components (for brevity we discuss them in terms of costs), based on the agent's internal acceptance of the rule (internal belief), its perception of societal support for the rule (perceived norms), and its belief about the purpose of the rule (rule-understanding). We use the example of an agent called Alex crossing an intersection when the light is red which is in violation of a traffic light abidance law when he takes a critically ill person to hospital:
 	\begin{itemize}
 	\item \textbf{Internal beliefs (B):} This indicates an agent’s subjective preference for rule interpretation (or rule’s objective/purpose). Breaking this imposes mental costs on an agent, whether or not others observe the action (inspired by \textit{behavioural beliefs} \cite{fishbein2011predicting}). Examples of such beliefs of Alex  include \emph{B.1:} all rules must be followed and \emph{B.2:}  this rule is because of the new mayor's non-sensical policies.
 	\item \textbf{The perceived norms (N):} This component indicates an agent's perception of societal support for the rule (e.g.~possible sanctions such as rebukes). Breaking these perceived norms imposes costs on an agent if other agents identify and sanction the violation (inspired by \textit{normative beliefs} \cite{fishbein2011predicting}). Alex may make different decisions under different situations based on his perceived norms, such as \emph{N.1:} when no one is around, and \emph{N.2:} when there are some of his neighbours all waiting for the light to turn green.
 	\item \textbf{Rule-understanding (C):} This component represents the rule as an agent understands it. This may differ from the real intention of the rule-maker. This is enforced by agents who have the duty of monitoring, reporting, and punishing the violators. This component has the most rigid punishments, such as dismissal, repaying the costs, and jailing. For executing this component, the system needs some official reports about the agents' behaviour (a formalised version of \textit{control beliefs} \cite{fishbein2011predicting}). Examples of theses include \emph{C.1:} there is no police around, \emph{C.2:} there is a working  traffic control camera, and \emph{C.3:} Alex may think he is allowed to cross the intersection the way that ambulances do (i.e.~he thinks he would not be punished by the police).
 \end{itemize}
 \item \textbf{Desires:} Agents have different desires with respect to the environment, their personalities, and the context, such as goals and ideal preferences. 
 \item \textbf{Personality:} Personality of an agent impacts different aspects, such as its learning method and the way it weights its beliefs. We use MBTI (Myers-Briggs Type Indicator) \cite{myers1998mbti} personality types. MBTI categorises personality types based on four dimensions: Introverted-Extraverted (I-E), Sensing-iNtuitive (S-N), Feeling-Thinking (F-T), and Judging-Perceiving (J-P). These dimensions are defined as follows:
 	\begin{itemize}
		\item Introverted versus Extraverted (I-E): This indicates where energy is oriented, or attitudes come from. 
		\item Sensing versus iNtuitive (S-N): This differentiates agents based on their \textit{Perceiving} methods (i.e.~collecting objective or abstract information).
		
		\item Feeling versus Thinking (F-T): This indicates how an agent makes decisions and \textit{judges} (i.e.~decisions are based on the demonstrable rationality or personal and social values).
		\item Judging versus Perceiving (J-P): This indicates how fast a person wants to reach a conclusion and `achieve closure'. Also, ``it describes the orientation to the extraverted world for every type''\cite{myers1998mbti}. 
	\end{itemize}
 These dimensions indicate the extreme points (i.e.~a person who is more Introverted is considered to be I).\footnote{In our model agents may have different values associated with different dimensions and this weight impacts their behaviour. Note that the weights are complementary (i.e.~10\% Introverted means 90\% Extraverted).} 
 
 Another method used for the same purpose, i.e.~to classify personalities, is called the \textit{Big Five} personality factors \cite{goldberg1990alternative}. The aspects (which are called traits by the designers) are measured by the model and its revised versions are: Agreeableness (the tendency to be cooperative or likeable), Conscientiousness (the extent to which an agent is achievement-oriented, dependable, and organised), Extraversion (how far an agent is socially-oriented, ambitious, and active), Openness (an agent's openness to experience and being unconventional), and Neuroticism (the inability of an agent to adjust to positive psychological states) \cite{Costaa}.

 \item \textbf{Skills and capabilities:} Some determinants of an agent's behaviour and the impacts of such behaviours are its skills and capabilities. 
 
 \item \textbf{Intention and decision:} An agent's intention to take an action and its decision about the final action is formed in this module (coloured in grey). The decision process includes normalising inputs based on aspects such as learnt parameters, personality, and roles, which results in an action. This action can be a modification of beliefs and roles or only performing a task. This module is at the core of the architecture and discussed in depth in the following section.
	\end{itemize}

	\subsection{Intention and decisions module}
	\label{decisionmodule}
 This module is at the core of the architecture and indicated by grey and dashed box in Figure \ref{architec}. The dashed lines in Figure \ref{architec} indicate all the relationships with the components now external to the decision module, including the visually conflated tripartite belief structure with its differentiated enforcement mechanisms explored previously. The components and moderators that integrate these in an agent’s final decision are discussed below:
	

	\begin{itemize}
 \item \textbf{Monitoring:} This block indicates an agent's belief about the chances of being punished by each of the three belief facets for a given violation. Note that the internal belief component does not need external monitoring mechanisms to work.
 \item \textbf{Punishments$ \mathbf{/} $rewards:} As discussed earlier, the punishments/rewards associated with internal beliefs and perceived norms depend on the agent’s beliefs. However, the rules generally have explicit and less ambiguous consequences than social norms in terms of punishment and rewards.
	\item \textbf{Modifiers:} As discussed earlier, an agent may decide about modifying its internal beliefs, role, and beliefs about the system for reasons such as cognitive dissonance (incurred costs for inconsistent internal beliefs \cite{aronson1999social}) and new information. This block thus functions as a moderator across the different influence factors.
	\item \textbf{Learning:} This block indicates an agent's interpretation of its observations, suggestions, and past information. Furthermore, the agent can increase its skills (including general skills) by observation or by practising over time.
	\item \textbf{Intentions and action decisions:} This block coordinates the decisions with cognitive or organisational consequences in terms of performing modifications or turning an intention into an action.
	\end{itemize}
 To have a better understanding of the way these modules work, let us explore those from a process perspective. First, an agent accounts for the consequences of an action with respect to \textit{monitoring} and \textit{punishment/rewards}. It also learns and updates its \textit{learning} of the system's characteristics with respect to its observations, and experience. Then given an agent's \textit{intention}, \textit{it assesses and revises its cognition to take an action}. If the agent decides to modify the beliefs and roles, it uses the \textit{modifier} for doing so. The agent also improves its cognition about system characteristics by employing a \textit{learning} procedure to improve and modify its understanding.

 In the following, we attempt to draw general links between weighing of beliefs and empirical data, so as to replicate diverse compliance behaviour as found in real-world settings.

	\subsection{Impact of personality}
	\label{person}
	 As depicted in Figure \ref{architec}, personality influences the decision module (i.e.~an agent's personality impacts its decision). Furthermore, the decision module is a mediator for modifying different cognitive aspects (e.g.~to resolve cognitive dissonance). In the following we discuss the impact of personality on an agent's utility function.

\textbf{Weighting perceived norms and internal beliefs:} To model the impacts of personality on norm conformance, we use the result of correlations between behavioural tendencies as measured by the \textit{California Psychological Inventory} (CPI) \cite{GoughCPI,TheCaliforniaPsychologicalInventory}. The measured aspect of interest for us is the classification of agents into the norm following vs. -questioning groups. The correlations between the CPI scales and the MBTI scores indicated that Sensing and Judging (S-J) in the MBTI personality types are correlated to norm-favouring personalities \cite{fleenor1997relationship}. Conversely, iNtuitive-Perceiving agents (N-P) appear to have the strongest tendency to question norms (i.e. weigh their internal beliefs higher than the perceived norms). 

\textbf{Dynamics in internal beliefs:} Perceived fairness of the system has disparate impacts on the agents' behaviour in terms of rule-following. The Thinking-Feeling (agreeableness for Big Five \cite{goldberg1990alternative}) aspect significantly influences agents' behaviour \cite{BuboltzJr2003,Skarlicki1999}. 
The impact of perceived fairness on ``\textit{organisational retaliatory behaviour}'' that was measured by several studies \cite{adams1965inequity,BuboltzJr2003,Skarlicki1999} indicate that an agent \textit{deliberately} changes its behaviour as a reaction to an unjust situation. 

\textbf{Mobility versus residency:} Residency in a place gradually leads to formation of friendship among agents; also, friendships have certain impacts on the agent's behaviour \cite{Argyle1984}. The Feeling types, weight keeping harmony in the society more \cite{myers1998mbti}. In our model, this effort can be modelled, based on the weights of the connections.

Having provided a high-level overview of the agent's cognition and decision-making modules, we describe an agent decision-making procedure using an example. For this purpose, we consider the condition under which an agent (A) observes that another agent (B) uses the company's properties for self-interest. First, we state how the \textit{Cognitive architecture} impacts agent A's decisions. Given the architecture presented in Figure \ref{architec}, the agent A's role impacts its decision. For example, if agent A is a manager 
of agent B reporting the action might be considered as a part of its responsibilities, while if it is a peer of agent B the same behaviour (i.e.~reporting agent B) might be considered as whistle-blowing. Such a change in role impacts agent A's beliefs about its actions --- i.e.~agent A may consider whistle-blowing the same as a betrayal to its friendship with agent B.

Also, agent A has different beliefs (i.e.~facets of belief) regarding whistle-blowing. For example, it may consider the rules too strict and considers the action a mild violation (internal belief) --- e.g.~agent B used company's copy machine to make copies of its own documents and the rule dictates firing of employees for such an action. Furthermore, agent A may have learnt that most of his colleagues consider whistle-blowing as a taboo (perceived norm). Finally, agent A interpretation of the rule may differ from the real intention of the company --- e.g.~company does not consider using a copy machine for making copies of a document as a violation.

Agent A's desires, personality, and skills and capabilities impact its decision. In the example stated earlier, maybe the agent A has certain desires for whistle-blowing \cite{ebader555692}, such as hatred of agent B. Agent A's personality impacts its desires, skills \cite{James2001}, and weighting its different belief facets \cite{fleenor1997relationship} (see Subsection \ref{person} for a clarification). Agent A's normalises its inputs in the intention and decisions resulting in the final action.

We use the stated example to describe an example to clarify how \textit{intention and decisions module} impacts an agent's final action. As indicated in Figure \ref{architec}, agent A's decisions are impacted by some external factors, including monitoring and punishment means. For instance, if agent A decides to report agent B's behaviour, it may change its mind to avoid social sanctions when the rest of it colleagues are present in the manager's room (i.e.~punishment and monitoring for perceived norm). Under the same circumstances, agent A may decide to report agent B, because its manager identified agent B's behaviour and there are strict punishments for collusion among agents (i.e.~punishment and monitoring for rules). 

To explain the learning and modifiers modules consider the earlier situation. Agent A learns the organisation's intentions of rules, its peers' expectations, and the punishments and rewards associated with a behaviour and modifies its earlier beliefs based on collected information and its observations. Also, there are situations such as agent A's loyalty to the organisation which bring it into the conflict between its peers' expectations. Therefore, agent A is afflicted with the cognitive dissonance that needs to be resolved by modifying its beliefs. This is exemplified by showing how smokers who were aware of the harms ranked their smoking habit milder than their counterparts \cite{TAGLIACOZZO1979393} (i.e.~changed their beliefs about their smoking behaviour). Note that this modification of beliefs impacts agents' future behaviours.

The stated example clarifies how different modules interact. However, there are instances such as the agency problem (known as the \textit{principal-agent problem}) where the monitoring cannot be performed easily \cite{Ross1973}. The principal-agent problem concerns the dilemma where the self-interested decisions of a party (agent) impact the benefits of the other person on whose behalf these decisions are made (principal) \cite{Mitnick2011}. Note that we consider utilitarian decisions made by agents who play an incomplete information game.\footnote{This approach was used for modelling agency problems \cite{Sedigh2019}.} In such games, agents learn the system's characteristics over time and use a utility function to make a decision. In the following section, we describe how we operationalise an agent's decision-making process by defining its utility function. 

\section{Operationalisation}
\label{opersec}
In this section, we concisely explain how an agent makes its decisions. We briefly describe an agent's utility function and the way it forms and updates its perception of norms. Note that an agent learns system's characteristics based on its personality,  
\subsection{Agent's decision utility function}
Agent $ A $ takes an action ($ x $) that maximises its utility function presented in Equation \ref{PunishFunc} ($ U_A (x) $). The utility function has four parts. The first part indicates the revenue that agent $ A $ earns for an action ($ R(x) $). The second part shows how the agent is mentally punished for such an action ($IBP_A(x)$), given its personality $ \big((N_A + P_A )/2\big) $.
\begin{equation}
\label{PunishFunc}
\begin{array}{@{}l@{}l@{}}
U_A (x)=&R(x)-\bigg(IBP_A(x) \times \dfrac{N_A + P_A}{2}\bigg)-\\
&\bigg((PN_A(x)\times NM_A)\times \big(1-\dfrac{N_A + P_A}{2}\big)\bigg)-\big(RP(x)\times  RM(x)\big)
\end{array}
\end{equation}

The third term shows the agent's perception of the punishment by its connections (social sanctions). Agent $ A $ estimates this, based on its perception of punishment for such an action $ (PN_A(x)) $, and it is moderated based on its estimation of its connections' monitoring strength $ (NM_A) $ and its personality $ \big(1-(N_A + P_A)/2\big) $. Finally, it takes account of organisational punishment regarding violation $ (RP(x)) $ and moderates its impact, based on its estimation of organisational monitoring $ (RM(x) )$. The next subsection discusses how agent $ A $ estimates monitoring strength associated with norms.

\subsection{Perception of norms}
The perception of norms differs from that of the rule by the vagueness of punishments associated with norms. To model an agent's perception of norms, we should note that a person expresses his/her beliefs, based on the weight he/she allocates to people's expectations. To consider this effect, we consider the impact of personality on an agent's norm conformance (i.e.~expectations with undefined consequences). 
\begin{equation}\label{NormLearning}
\begin{array}{@{}l@{}l@{}}
{ PN_A^t=wi\times \big(\dfrac{\sum_{i\in C_{A,N}}^{}EN_i^t}{K_{A}}+\dfrac{(K_{A}-K_{A,N})\times PN_i^{t-1}}{K_{A}}\big)+ (1-wi)\times PN_A^{t-1}}
\end{array}
\end{equation}

Equation \ref{NormLearning} shows how agent $ A$ (agent of interest) collects its associated agents' opinions about the norm (they may express their opinions different than their real internal beliefs), and updates the society's expectations from itself at time $ t $ $\big(PN_A^t\big)$ (in former formula we skipped $t$ for simplicity). The agent associates some weights with the recent information (i.e.~$ wi $). Furthermore, the agent averages the recommended scores by its connections with whom it has strong relationships (i.e.~the associated $ wi\geq 0.5$). From those connections, it takes account of the ones who have at least the same experience as itself; we call the subset of such members from $ C_A $ (i.e.~the whole connections of $ A $) as $ C_{A, N} $. Also, $ K_A $ and $ K_{A,N} $ indicate the number of members of $ C_{A} $ and $ C_{A,N} $, respectively. This procedure states that an agent would not ask for less experienced agents' recommendations because of the feeling that it knows better. Furthermore, an agent does not criticise people's expectations unless it knows the audience well (i.e.~the ones who have a higher $ w_{iA} $). In addition, agent $ A $ assumes that the rest of the connections (i.e.~$ K_A-K_{A,N} $) realise the norm the same as himself $\big(PN_i^{t-1}\big) $. Then the agent associates the rest of the weights (i.e.~$ 1-wi $) with its past perception about expectations (i.e.~$ PN_A^{t-1} $).

Now we discuss how other agents express their beliefs to agent $ A $. Agent $ i $ does it through a weighting of its perception of social expectations ($ PN_i $) and its beliefs ($ IB_i $). Equation \ref{NormExpress} shows how weighting the outer world impacts an agent's expression of beliefs. As discussed earlier, the personality of agent $ i $ (a friend of agent $ A $) impacts its expression of its beliefs about the outer world's expectations. $ EN_i^t $ indicates agent $ i $'s expression of what people do and what they should do at time $ t $. As discussed earlier, the iNtuitive and Perceiving aspect of an agent decreases weights given to other people's expectations; hence, agent $ i $ straightforwardly expresses its own beliefs and expectations at that time (i.e.~$ IB_i^t $, the first term on the right of Equation \ref{NormExpress}). The second term on the right shows the opposite --- i.e.~how personality of agent $ i $ impacts its expression of ideas, based on the value it associates with other agent's expectations (i.e.~$ PN_i^t $).
\begin{equation}\label{NormExpress}
EN_i^t=\big(IB_i^t\times \dfrac{N_i + P_i}{2}\big)+\big(PN_i^t\times (1-\dfrac{N_i + P_i}{2})\big)
\end{equation}

\section{Simulation, algorithms, and parameters}
\label{SimulationAndPArameters}
In this section, first, we discuss the underlying assumptions of the simulation. Then we provide an overview of two historical societies studied for simulation, namely the English East India Company (EIC) and Armenian merchants of New-Julfa (Julfa). Then we briefly discuss their aspects of interest for us and the simulation procedures used to represent agents' behaviour in these societies.
\subsection{Assumptions}
This paper is inspired by empirical studies on the importance and effects of cognitive dissonance --- tensions formed by conflicts between different cognitions \cite{aronson1999social} --- on agents' behaviour. This leads to creating some justification for taking one of the conflicting actions. Empirical studies attributed workers' productivity to cognitive dissonance formed by fairness of institutions \cite{Adams1962}. Studies showed that procedural justice (having fair dispute resolution mechanisms) \cite{Sunshine2003}, and payment schemes (e.g.~underpaying or overpaying)  \cite{adams1965inequity} impact agents' behaviour. Also, as discussed earlier, the impact of fairness of institutions on agents' behaviour varies based on their personalities.

\subsection{Societies}
\label{socielab}
  \textbf{Armenian Merchants of New-Julfa ({Julfa}):} Armenian merchants of New-Julfa were originally from old Julfa in Armenia. They re-established a trader society in New-Julfa (near Isfahan, Iran) after their forced displacement in the early 17{th} century \cite{Herzig1991}. 
   They used commenda contracts (profit-sharing contracts) in the society and also used courts to resolve disputes. The mercantile agents were responsible for buying and selling items and moved among different nodes of the trading network expanded from Tibet (China) and Manila (Philippines) to Marseille (France) and  Venice (Italy) \cite{aslanian2007indian}.
   
   
   \textbf{The English East India Company (EIC):} During the same time, the English contemporaneous counterpart of Julfa (i.e.~the English \textit{East India Company} (EIC, AD 1600s-1850s)) had a totally different perspective on managing the society. They built settlements for their mercantile agents to stay in India \cite{hejeebu2000microeconomic}. Also, they paid their employees fixed wages and fired agents based on their beliefs about their trading behaviour \cite{hejeebu2005contract}.
   
   Note that the EIC's trading period covers two events, namely a) granting permission for private trade (1665-1669) and b) a significant budget deficit on part of employees (around 1700). The EIC reduced its employees wages when granted the permission for private trade (i.e.~fairness of the system had decreased because agents' already performed private trade without such a permission). Furthermore, agents were desperate for their living costs because of the budget deficit\footnote{\label{neweastfoot}This deficit was because of lower wages and profits and higher living costs due to an increase in the number of private traders as a consequence of formation of the New East India Company (see p.~17 of \cite{marshall1976east}).}. In the simulation model, we regenerate those events in iterations 70 and 100.

          \textbf{Fairness:} Another difference between the two historical long-distance trading societies is associated with their payment schemes and adjudication processes. The EIC rarely employed an adjudication process, and they paid low wages to their employees \cite{hejeebu2005contract}. However, in Julfa a mercantile agent was paid based on his performance \cite{aslanian2007indian}. Julfans had adjudication processes to resolve disputes considering available evidence \cite{aslanian2007indian}. However, the Julfa society had certain characteristics that questioned its complete fairness --- for instance, in the Julfa society, the capital of the family was managed by the elder brother\footnote{This rule helped the families to work like a firm.} \cite{Herzig1991}. This rule deprived younger brothers of managing their own capital.
   
      \textbf{Agent's mobility:} Another difference in the policy of the two historical long-distance trading societies concerns whether or not the mercantile agents stayed in a certain place to perform trades. In the EIC, the company built its own factories and mercantile agents (but not sailors) resided there \cite{hejeebu2000microeconomic}. This introduces issues such as forming an informal community within the company that could have operated based on the norms of friendship \cite{chaudhuri1965english}.
   
   Note that the events of years 70 and 100 are regenerated for unfair societies.\footnote{Some discussions suggest that agents tend to break bureaucratic rules that seemed to be harmless for organisation \cite{erikson2014between}.} Note that once private trade is permitted, the fairness of the society decreases (because of a decrease in wages).

          	\subsection{Algorithms}
 In this section, we discuss the procedures employed to use the above-described model to simulate the two aforementioned societies. The simulation model is split into four distinctive procedures and one sub-procedure. The first procedure is the meta-algorithm that executes other algorithms in an appropriate sequence and updates parameters required for them. The second procedure models the societal level activities of the simulation, including creating an initial population and staffing (hiring new recruits) to create a stable population. The fourth procedure models \textit{mercantile agents}' decision-making and learning the system's parameters. The fourth procedure also includes a sub-procedure for defining new parameters associated with hired mercantile agents. The fourth procedure covers the decision-making and learning procedure associated with managers (i.e.~monitoring agents).

	\begin{algorithm}[hbt!]
	\caption{Meta-algorithm}
	\DontPrintSemicolon
	\label{meta}
	\SetKwBlock{Begin}{}{}
	\tcc{Intialise the system starting with $ iteration \gets 0 $.}
	Create 500 new agents with $ status \gets new $, random personality aspects, and random parameters\;
	Assign appropriate roles (i.e.~mercantile, managers, and directors) to created agents\;
	\tcc{Call algorithms in an appropriate sequence.}
 \Repeat{$ iteration=250 $}{
		Run Algorithm \ref{Initiatealgo}\;
	Run Algorithm \ref{factoralgo}\;
	Run Algorithm \ref{MidManager}\;
			\lIf{Private trade is legalised and $ iteration = 100 $}
		{
			Mercantile agents feel desperate for basic living costs
		}
	
	$ iteration \gets iteration + 1  $\;
}
 
\end{algorithm}

    \textbf{Algorithm \ref{meta}} shows how the simulation is initialised, other procedures are executed, and the system's characteristics are modified. In iteration 0, the system is initialised by creating 500 new agents with random parameters (line 1). The roles are assigned to created agents (about 2\% directors, 5\% managers, and the rest mercantile agents)\footnote{These numbers are inspired from the numbers in the EIC \cite{hejeebu2000microeconomic}.}, and they have 0 years of experience (line 2). The rest of the algorithm is repeated until the termination condition is met (lines 4-9). 
	
		\begin{algorithm}[hbt!]
		\caption{Societal level set-up}
		\DontPrintSemicolon
		\label{Initiatealgo}
			\tcc{$ n $ equals deceased and fired agents (mercantile agents and managers) in the former iteration.}
			The most experienced mercantile agents get promoted to a managerial role to keep the number of managers constant.\;
		Create $ n $  new agents $ status \gets new $, $ Experiene \gets 0 $, randomly initialise personality aspects and other parameters (discussed in Table \ref{paramet3}) \;
	\end{algorithm}
    In each iteration, the simulation begins with the societal algorithm (i.e.~Algorithm \ref{Initiatealgo}, line 4). Then the algorithm associated with the mercantile agents is run (i.e.~Algorithm \ref{factoralgo}). Note that the Algorithm \ref{initialise} associated with the newly recruited mercantile agents is called inside Algorithm \ref{factoralgo}. After 100 years, in unfair system, private trades are legalised and the mercantile agents face financial issues and feel financially desperate. This phenomenon was observed in the EIC after a decrease in wages, coupled with fewer opportunities for private trade because of the establishment of the New East India Company. Note that for simplicity, when we modelled the phenomenon, we did not consider a gradual decrease (line 6). Finally, the simulation iteration increases by one, and the stop condition is checked (line 8).

	\textbf{Algorithm \ref{Initiatealgo}} shows how the societal level of the system is simulated. The organisation hires and promotes agents to sustain the number of agents per role --- i.e.~by replacing deceased agents (lines 1-2).

\begin{algorithm}[hbt!]
		\caption{Initialising the mercantile agent's algorithm}
		\DontPrintSemicolon
		\label{initialise}
		\SetKwBlock{Begin}{}{}
	
			$Fair \gets Random Uniform (0,1)$\;
			\tcc{Rule-understanding and the internal belief about rules.}
Agent has a random perception of norms and rules\;	
$ Status\gets Experienced $\;
	\end{algorithm}

    \textbf{Algorithm \ref{initialise}} shows how the parameters of newly introduced mercantile agents (i.e.~new agents) are initialised. This algorithm is executed only for inexperienced agents (i.e.~new recruits). An agent has a completely random understanding of the system's characteristics (lines 1-2). After initialisation, the agent updates its status to \textit{experienced} so that the system can identify it (line 3).

	\begin{algorithm}[hbt!]
	\caption{Mercantile agent's algorithm}
	\DontPrintSemicolon
	\label{factoralgo}
	\SetKwBlock{Begin}{}{}
	\tcc{Upadate parameters for new recruits.}
	\lIf{Status = New }
	{Set agent's parameter using Algorithm \ref{initialise}}
\If{$ Experience > 3 $}{
		\tcc{Make a decision on cheating using Equation \ref{PunishFunc}.}
	$ Cheat \gets 0$\;
		\tcc{$\mathcal Viol $ is a set of random violations.}		
	\ForEach {$viol \in \mathcal Viol $}{
	 $util \gets Utility(viol)$\;
	 \lIf{$Util>Utility(Cheat)$}{$Cheat\gets viol$ }
}

}

\If{Desperate}{
	\ForEach {Violation level}{
		\tcc{Punishment of internal beliefs decreases significantly}
		Change the costs of internal belief violations 
	}

}

	\tcc{Agent ($ I $) increases the weight of fellows.}
$ W_{Ij}\gets W_{Ij}\times (1 + \#Rnd(FriendIncrease)) \ \forall_{j\in frinds}$\;

\If{Moving to a new place}
{\tcc{Most of the former fellows are replaced with new fellows}
	Replace Mob\% of fellows with new fellows.
}


	Learn system parameters and modify your internal belief and perceived norm\; 

	\tcc{Agent may die}
	$ Experience \gets Experience + 1 $\;
	\lIf{$Rand(1) \leq MortalityProbability(Experience + 15) $}{Die}
\end{algorithm}

    \textbf{Algorithm \ref{factoralgo}} shows the procedures associated with mercantile agents' cognition and decision-making processes. Note that in this algorithm $ \#Rnd(x) $ indicates a random number generated in the interval $ (0,x) $. As stated earlier, if the status of the mercantile agent is new, he goes through an initialisation algorithm (i.e.~Algorithm \ref{initialise}, line 1). Experienced mercantile agents decide whether to cheat based on a set of potential violations (lines 2-6). Lines 7-9 model the scenario when mercantile agents faced a decrease in wages and a drop in private trade's revenue such that they were desperate to pay for their living costs. As a consequence, costs associated with violating the internal beliefs decrease (lines 7-9). Also, the mercantile agent increases the weight of social bonds with his associated agents (line 10). And when moving to a new place, the majority of a mobile mercantile agent's work fellows (e.g.~Julfans) are replaced (lines 11-12). Finally, the mercantile agent learns system parameters, modifies his beliefs, increases his experience, and dies with an estimated probability (lines 13-14).

	\begin{algorithm}[hbt!]
	\caption{Manager's algorithm}
	\DontPrintSemicolon
	\label{MidManager}
	\tcc{Manager reports (and eventually punishes) a number of employees who violate the rules of the organisation beyond its tolerance level.}
	$ PotPunish\gets$ employees with violations more than $ TolPunish $\;
\eIf{The number of members of $ PotPunish > MaxPunish$}{
	\tcc{The manager has a limit for the number of agents he can punish called $ MaxPunish $.}
Punish $ MaxPunish $ out of $ PotPunish $ that have the most violation
}
{
	Punish all $ PotPunish $ members.
} 	
$ Experience \gets Experience + 1 $\;
\lIf{$Rand(1) \leq MortalityProbability(Experience + 15) $}{Die}
\end{algorithm}	\setlength{\floatsep}{0.1cm}

    \textbf{Algorithm \ref{MidManager}} shows the procedures associated with managers. A manager creates a set that consists of violators with unacceptable violations (i.e.~it tolerates violations to some extent, see line 1). Note that the manager reports about the violators and punishes a certain number. If the number of violators exceeds a certain threshold, it punishes the worst violators (lines 2-3). Otherwise, all the violators are punished (lines 4-5). Finally, the agent's experience and age increase, and the agent may die (lines 6-7).

\subsection{Simulation parameters}
\label{subs:param}
In this subsection, we discuss the important parameters employed in the simulation (see Table \ref{paramet3}), along with the justification for the parameterisation. Note that we used 250 iterations to reflect the longevity of the EIC (it was active with some interruptions and changes in power from 1600 to 1850). Each iteration thus models one year of activity in these systems. In Table \ref{paramet3}, column `Name' indicates the name of parameters, column `Comment' shows additional information if required, column `Distribution' indicates the probability distribution of these parameters, and column `Values' indicates the values of parameters estimated for the two historical long-distance trading societies. Note that these parameters can be easily revised to reflect other societies.

\begin{table}[htb!]
\center
					\caption{Parameters associated with the model}
\label{paramet3}
	\begin{tabular}{|l|l|l|l|}

		\hline	\multicolumn{1}{|c|}{Name} & \multicolumn{1}{c|}{Comment} & \multicolumn{1}{c|}{Distribution} & \multicolumn{1}{c|}{Values} \\ \hline
		\multicolumn{1}{|l|}{Fairness} & \multicolumn{1}{l|}{Unfair : Fair} & \multicolumn{1}{l|}{Constant} & \multicolumn{1}{l|}{$-0.4:0.6$} \\ \hline
		Perception of fairness of system &  &  $Uniform$ & $(-1,1)$ \\ \hline
		\multicolumn{1}{|l|}{Rule monitoring} & \multicolumn{1}{l|}{\begin{tabular}[c]{@{}l@{}}The lower bound $(lb)$\\The learnt probabilities\end{tabular}} & \multicolumn{1}{l|}{$Uniform$} & \multicolumn{1}{l|}{\begin{tabular}[c]{@{}l@{}}$(0,1)$\\$(lb,1)$\end{tabular}} \\ \hline
		\multicolumn{1}{|l|}{Norm monitoring strength} & \multicolumn{1}{l|}{} & \multicolumn{1}{l|}{$Uniform$} & \multicolumn{1}{l|}{$(0,1)$} \\ \hline
		\multicolumn{1}{|l|}{Minor violation punishment} & \multicolumn{1}{l|}{Probability} & \multicolumn{1}{l|}{$Uniform$} & \multicolumn{1}{l|}{$(0,1)$} \\ \hline
		\multicolumn{1}{|l|}{Weight of the connections} &  Fellows &  $Uniform$ &  $(0,1)$ \\ \hline
\multicolumn{1}{|l|}{Modifier for mental costs} & \multicolumn{1}{l|}{Cognitive dissonance} & \multicolumn{1}{l|}{$Uniform$} & \multicolumn{1}{l|}{$ (0, 0.3) $} \\ \hline

\multicolumn{1}{|l|}{Modifier for connections} & \multicolumn{1}{l|}{Friendship} & $Uniform$ & \multicolumn{1}{l|}{$ (0, 0.2) $}  \\ \hline
\end{tabular}
\end{table}



\textbf{Fairness:} Note that as stated earlier, Julfa had fairer institutions than the EIC due to profit sharing and using adjudication processes. We set system fairness for fair and unfair societies as 0.6 and -0.4, respectively. We believe that neither of these two societies was completely fair or unfair (e.g.~EIC managers justified the firing of agents in their letters, which indicates some efforts towards interactional fairness \cite{hejeebu2005contract}).

\textbf{Perception of fairness:} An inexperienced agent has a completely random understanding of system characteristics.

\textbf{Rule and norm monitoring:} The agents believe the more serious violations are more likely to be punished. They use a Uniform probability distribution that is identified by its lower and upper bounds. For the upper bound, they use 1 (i.e.~completely violating the rule such as stealing all the money). However, they are unsure about the lower bound for tolerance of peers and managers; hence, a continuous Uniform probability distribution in (0,1) is used. For the strength of the norm monitoring (i.e.~how often agents punish each other), agents have a random understanding. However, these parameters are updated by continuous learning.


\textbf{Minor violation punishment}: In the model, we assume the agents have doubts about being fired for minor violations. This doubt is modelled by using uniform random numbers.

\textbf{Weight of the connections:} To model all the possibilities for new links, such as knowing one another in advance, we used a random number (0,1) for new friendship links and links between mercantile agents and middle managers.

\textbf{Modifier for mental costs:} When an agent violates a rule, it randomly discounts the costs associated with such an action by a maximum of $ 10\% $ of its initial point and on average the initial costs decreases by $ 5\% $ of its initial point.

\textbf{Modifier for weights of connections}: For changes in weights of connections, we increase the connection's weight as a proportion of the current weight (i.e.~$weight \times x$). Due to the ongoing interactions, the weight of friendships increases randomly between (0, 0.2).

\textbf{Learning}: Furthermore, we parametrise the agents' learning as follows. Agents discount past information using a weight of 30\%  (i.e.~a weight of 70\% for recent information). This reflects the importance of recent information for agents, because they are not sure about the stability of the system's behaviour in the long run. Also, we use entrepreneurs' personalities as a representative of personality of these trading agents  \cite{Zhao2010} (see Table \ref{entre}). 
\begin{table}[hbt!]
	\caption[Tendency of personalities to be entrepreneurs]{Tendency of personalities to be entrepreneurs and the impact of them on an organisation's performance}
	\label{entre}
	\centering
	\begin{tabular}{|C{2.8cm}|C{2.8cm}|C{2.8cm}|C{2.8cm}|}
		\hline
		\textbf{Big Five$^1$} & \textbf{MBTI$^{2}$} & \textbf{Intention$ ^3 $} & \textbf{Performance$ ^4$} \\ \hline
		\textit{E} & \textit{E} & 0.11 & 0.05 \\ \hline
		\textit{O} & \textit{N} & 0.22 & 0.21 \\ \hline
		\textit{A} & \textit{F} & -0.09 & -0.06 \\ \hline
		\textit{C} & \textit{J} & 0.18 & 0.19 \\ \hline
		\multicolumn{4}{l}{\begin{tabular}[c]{@{}l@{}}$^1$\textbf{E:} extraversion, \textbf{O:} openness to experience,  \textbf{A:} agreeableness, \textbf{C:} conscientiousness.\end{tabular}}\\
		\multicolumn{4}{l}{\begin{tabular}[c]{@{}l@{}}$^{2}$\textbf{E-I:} Extravert-Introvert dimensions, \textbf{S-N:} Sensing-iNtuitive dimensions, \textbf{T-F:} \\Thinking-Feeling dimensions, \textbf{J-P:} Judging-Perceiving dimensions. Note that initials\\show the aspect used --- for instance, E is the degree to which the person has the\\ Extravert aspect. Note that as with their earlier study, we assume that results of\\ different tests are interchangeable (see Appendix of \cite{Zhao2006}). \end{tabular}}\\
		\multicolumn{4}{l}{\begin{tabular}[c]{@{}l@{}}$^3$It shows the correlation between personality and the \textit{intention} of the person to be\\ an entrepreneur. \end{tabular}}\\
		\multicolumn{4}{l}{\begin{tabular}[c]{@{}l@{}}$^4$It shows the correlation between personality and the \textit{performance} of the person\\ as an entrepreneur. \end{tabular}}
	\end{tabular}
\end{table}

\section{Results}
\label{results}
In this section, we discuss the simulation results considering four different combinations of two characteristics, namely a) mobility of agents and b) fairness of institutions.

We utilised \textit{NetLogo} \cite{netlogo} to perform our simulation. We also used 30 different runs for each set-up and then averaged their results. Table \ref{specdyn} indicates the characteristics for the \textit{four} simulated societies and societies they represent. The set-ups (i.e.~societies) are identified by the first letter of the characteristics, namely \textit{M} and \textit{F} that are representatives of the \textbf{m}obility of agents across trading nodes (M) and \textbf{f}airness of the institutions (F), respectively. We used a Boolean index to indicate whether such an attribute is included (i.e.~1) or not (i.e.~0). Likewise, in the table, a tick indicates that the society possesses such an attribute, and a cross indicates the society does not possess such an attribute.  In the absence of detailed knowledge, we assume an explorative approach that starts off with a prototypical EIC configuration, and incrementally approximates the Julfa configuration by hypothesised intermittent society configurations. This approach allows for a nuanced analysis of the individual influence factors, so as to isolate their respective influence on the simulation outcome. Table \ref{specdyn} reflects this by gradually modifying individual characteristics of the EIC ($M_0F_0$) society towards the Julfa ($M_1F_1 $). In the following figures, we use two vertical dashed-lines that indicate the year that permissions for private trade was granted and the year that agents face issues with their living costs (i.e.~year 70 and 100, respectively).


\begin{table}[hbt!]
	\centering
	\caption{System specification based on different characteristics}
	\label{specdyn}
	\begin{tabular}{lcccc}
		\hline
		Characteristics & {\begin{tabular}[c]{@{}c@{}}$M_0F_0$\\ (EIC)\end{tabular}} & {$M_0F_1 $} & {$M_1F_0 $} &
		   
		  			{\begin{tabular}[c]{@{}c@{}}$M_1F_1 $\\ (Julfa)\end{tabular}} \\\hline
			Mobility & \ding{55} & \ding{55} & \ding{51} & \ding{51} \\
			Fairness & \ding{55} & \ding{51} &  \ding{55} & \ding{51} \\\hline
		
	\end{tabular}
\end{table}

\begin{figure}[hbt!]
	\centering
	\includegraphics[trim=0cm 2.5cm 0cm 0cm,clip,width=1\textwidth]{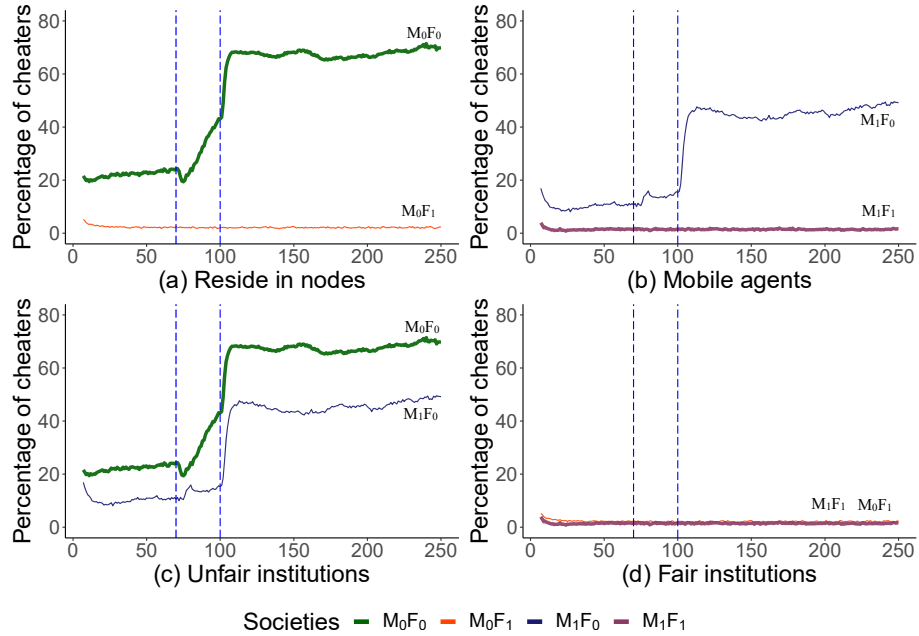}
	\caption[Cheating frequency (A)]{Cheating frequency}
	\label{Cheater16A}
\end{figure}

\begin{figure}[hbt!]
	\centering
	\includegraphics[trim=0cm 2.5cm 0cm 0cm,clip,width=1\textwidth]{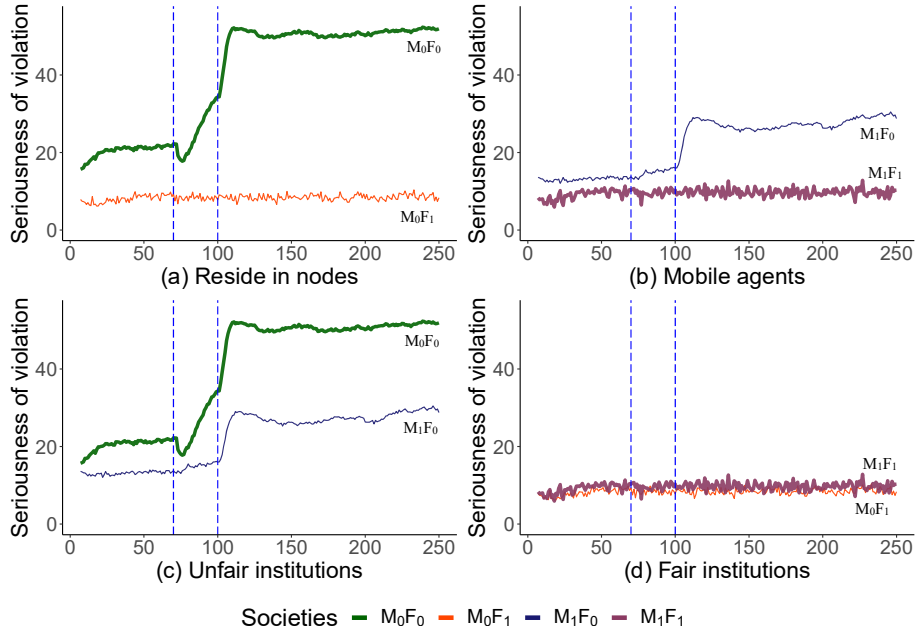}
	\caption[The seriousness of violations (A)]{The seriousness of violations }
	\label{MeanCheat16A}
\end{figure}

Certain characteristics of societies impact agents' tendencies to violate rules. Figures \ref{Cheater16A}a-\ref{Cheater16A}d present the percentage of mercantile agents who violated the rules (i.e.~cheaters) in a year. In these plots, the y-axis represents the percentage of cheaters. As can be seen, the most influential characteristic is the fairness of institutions (Figure \ref{Cheater16A}c versus \ref{Cheater16A}d) that reduces the number of cheaters significantly. Also, mobility of agents (Figure \ref{Cheater16A}a versus \ref{Cheater16A}b) reduces the number of cheaters moderately. Note that as shown in Figure \ref{MeanCheat16A}d, even in fair societies some cheaters are available; however, this number is not noticeable because of high percentages of cheaters in other configurations. Also changes in fairness of institutions (first dashed-line) and financial issues that lead to an increase in the percentage of cheaters.

Another important issue regarding cheating is the seriousness of a violation. Figures \ref{MeanCheat16A}a-\ref{MeanCheat16A}d present results for this phenomenon. In these plots the y-axis indicates the average of the seriousness of violations of cheaters. As can be seen, societies with fair institutions and the mobility of the agents had less serious violations in the long-run. In other words, Figures \ref{MeanCheat16A}a-\ref{MeanCheat16A}d indicate relatively the same patterns with respect to the percentage of cheaters shown in Figures \ref{Cheater16A}a-\ref{Cheater16A}d. 

 Note that the combination of these results indicates that, in a society with unfair institutions where the agents reside in a node (e.g.~the EIC), a higher percentage of agents cheat and the cheaters commit more severe frauds (i.e.~frauds with more costs for the company). Our model mirrors the results what was observed in the EIC. For instance, the real EIC situation was much worse than that of our simulation results. The following case is an example that indicates another popular cheating mechanism (i.e.~embezzlement) and its extent in the system:

\begin{displaycquote}[p.~466]{chaudhuri2006trading}
``The most common practice of partial defraudment 
in the Indies was to enter large sums of money in the name of fictitious
Asian merchants as advance payment for goods and use the money to finance
the private trade of the servants.''
\end{displaycquote}

As Aslanian points out, the ``cases of cheating and dishonesty are rarely mentioned in Julfan correspondence'' \cite[p.~249]{aslanian2007indian} . Furthermore, based on the historical evidence, he notes that blacklisting in Julfa was extended to cases such as refusing to pay the share of taxes \cite[p.~249, footnote 66]{aslanian2007indian}. This indicates the honesty of Julfan traders that made it possible to boycott society members for reasons other than violating trade rules.

Another instance concerns the historical evidence from the chartered companies (e.g.~EIC) and Julfa to evaluate the patterns suggested by our simulation results. The evidence is presented in the form of quotations from historians. The passage below belongs to the EIC:

 \begin{displaycquote}[p.~325]{chaudhuri2006trading}
``There were even fraudulent attempts to charge the Company [EIC] a higher price by buying it [pepper] during the cheap season and then entering
in the books the later price which had been raised by the demand from private traders arriving late on the coast. In fact, the Court of Directors
felt so strongly about the expenses incurred at Tellicherry [sic] under the management of Robert Adams that they were prepared to abandon the
settlement altogether unless the charges were drastically reduced.''
 \end{displaycquote} 
Chaudhuri notes that the reason for the court's decision was a downswing in the pepper market inside Europe \cite{chaudhuri2006trading}. However, this argument not only suggests the cheating behaviour and acceptance of it, but it also points to the beliefs about some coalitions. 
\section{Discussion, concluding remarks, and future directions}
\label{conc}

This study has presented a conceptual agent cognitive architecture inspired from the BDI \cite{Bratman1988} and TPB \cite{fishbein2011predicting} models to investigate the interactions of different belief facets with institutions. More precisely, in this study we have used the idea of different belief facets \cite{fishbein2011predicting}, along with their monitoring characteristics, to model an agent's decision-making process. Finally, in this study, we have used the evidence from empirical studies to apply the cognitive architecture to the context in a comparative simulation model of two historical long-distance trading societies, namely Armenian merchants of new Julfa (Julfa) and the English East India Company (EIC).

For simulating the aforementioned historical cases, we have considered two characteristics of these systems, namely mobility of agents and fairness of institutions, which are central features for cooperation in long-distance trading. The simulation results mirror the historical evidence for these two societies. The results show that the fairness of institutions is a pivotal characteristic to deter agents from cheating. Moreover, in a fair institutional setting, non-compliant agents cheated with less severity (i.e.~imposed lower costs on the system). Also, the increase in living costs (in iteration 100 --- second vertical line), that led to an increase in percentage of cheaters and the seriousness of cheats was because of a decrease in costs associated with breaking internal beliefs (see Section \ref{socielab}, footnote \ref{neweastfoot} and Algorithm \ref{factoralgo}, lines 7-9). This result emphasised the importance of this aspect of agents' cognition (i.e.~internal beliefs) in following the rules. 

Finally, as shown in Figures \ref{Cheater16A} and \ref{MeanCheat16A} ($M_1F_0$), mobility of agents helps the organisation by 
lowering the inclination of agents to break rules. This is caused by higher mobility, and thus fewer opportunities for agents to form strong friendship bonds and express their real beliefs to each other; leading to overall higher levels of compliance. The impact of fairness of institutions on agents’ behaviour translates well in modern settings; in organisational studies it is referred to as organisational citizenship (i.e.~cooperative working environment) and counterproductive work behaviour (i.e.~agents deliberately decrease their cooperation) \cite{Spector2010}. The impact of internal beliefs (e.g.~religions) on agents' behaviour has also been observed by political philosophers such as Machiavelli.\footnote{Machiavelli discussed internal beliefs could help a governor to bring order into society: ``[T]hose citizens whom the love of fatherland and
		its laws did not keep in Italy were kept there by an oath that they were forced to
		take; [... t]his arose from nothing other than that religion Numa [sic] had introduced in that city''(p.~34 of \cite{mansfield1998discourses}).}


As a future extension of the current study, detailed interactions between other modules of the cognitive architecture presented in Figure \ref{architec} deserves more attention. The societal model can be extended to take account of a wider range of characteristics such as apprenticeship programmes, and considering the impact of those on organisational profitability so as to sketch a more realistic picture of these historical trading societies. 
%

%
%
%
 \bibliographystyle{splncs04}
\bibliography{ref}

\end{document}